\documentclass[letterpaper, 10 pt, conference]{ieeeconf}  

\IEEEoverridecommandlockouts                              

\usepackage{amsmath,amssymb,bm,mathtools}

\usepackage{algorithm}

\usepackage{algorithmicx}
\usepackage{algpseudocode}

\usepackage{hyperref}

\usepackage{booktabs}
\usepackage{amssymb}

\title{\LARGE \bf
Push, Press, Slide: Mode-Aware Planar Contact Manipulation via Reduced-Order Models}

\author{Melih Özcan$^{1}$,  Umut Orguner$^{1}$, and Ozgur S. Oguz$^{2}$
\thanks{$^{1}$Melih Özcan and Umut Orguner are with the Dept. of Electrical and Electronics Engineering, Middle East Technical University, Ankara, Turkiye.
        {\tt\small \{melih.ozcan, umut\}@metu.edu.tr}}%
\thanks{$^{2}$Ozgur S. Oguz is with the Dept. of Computer Engineering, Bilkent University, Ankara, Turkiye.
        {\tt\small ozgur@cs.bilkent.edu.tr}}%
}

\begin{document}

\maketitle
\thispagestyle{empty}
\pagestyle{empty}

\begin{abstract}

Non-prehensile planar manipulation, including pushing and press-and-slide, is critical for diverse robotic tasks, but notoriously challenging due to hybrid contact mechanics, under-actuation, and asymmetric friction limits that traditionally necessitate computationally expensive iterative control. In this paper, we propose a mode-aware framework for planar manipulation with one or two robotic arms based on contact topology selection and reduced-order kinematic modeling. Our core insight is that complex wrench-twist limit surface mechanics can be abstracted into a discrete library of physically intuitive models. We systematically map various single-arm and bimanual contact topologies to simple non-holonomic formulations, e.g. unicycle for simplified press-and-slide motion. By anchoring trajectory generation to these reduced-order models, our framework computes the required object wrench and distributes feasible, friction-bounded contact forces via a direct algebraic allocator. We incorporate manipulator kinematics to ensure long-horizon feasibility and demonstrate our fast, optimization-free approach in simulation across diverse single-arm and bimanual manipulation tasks. Supplementary videos and additional information are available at \url{https://sites.google.com/view/pushpressslide}.
\end{abstract}

\section{INTRODUCTION}
\label{sec:introduction}

\begin{figure}[htb]
    \centering
    \includegraphics[width=0.48\columnwidth]{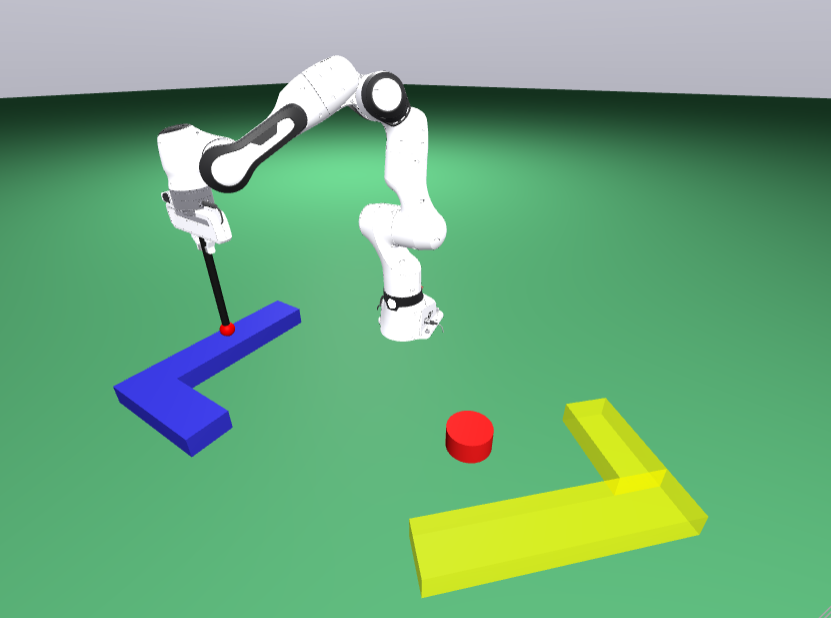}\hfill
    \includegraphics[width=0.48\columnwidth]{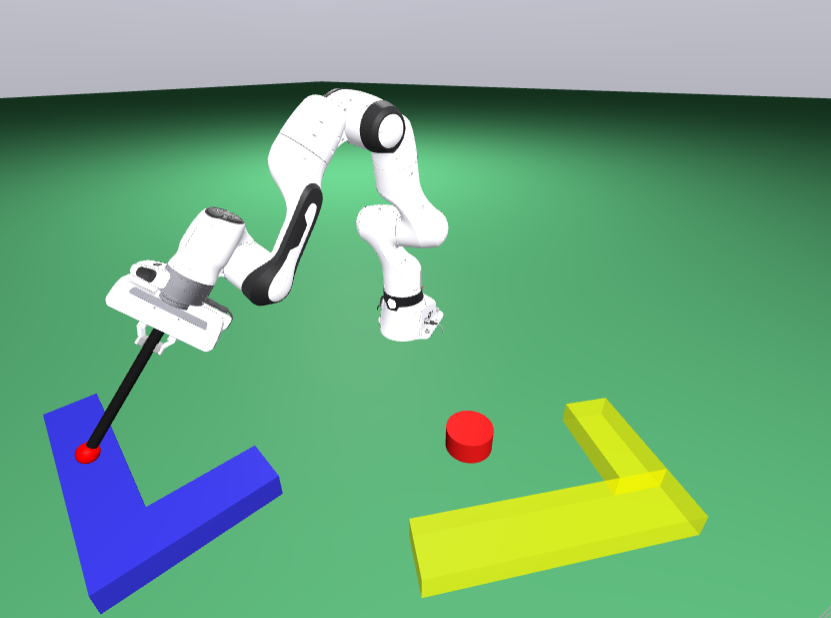}\\[1ex]
    \includegraphics[width=0.48\columnwidth]{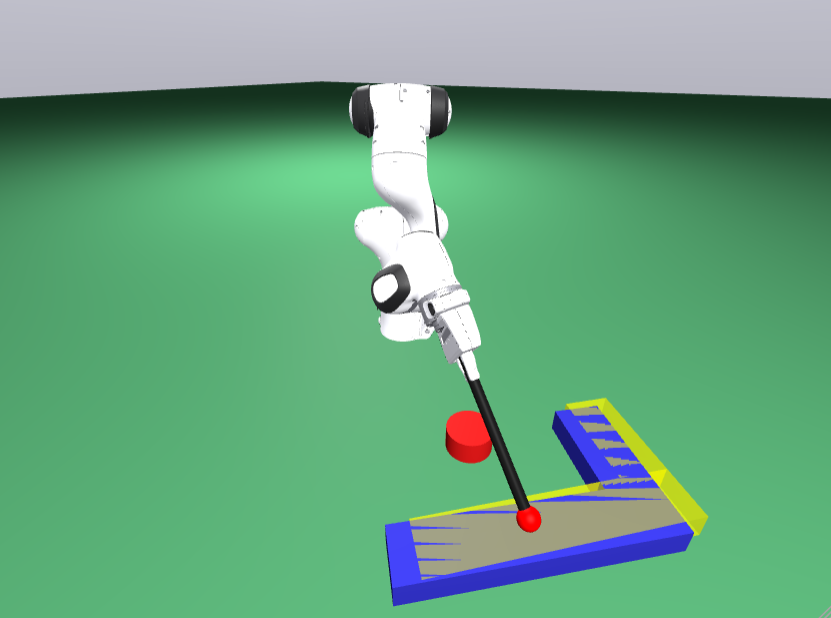}\hfill
    \includegraphics[width=0.48\columnwidth]{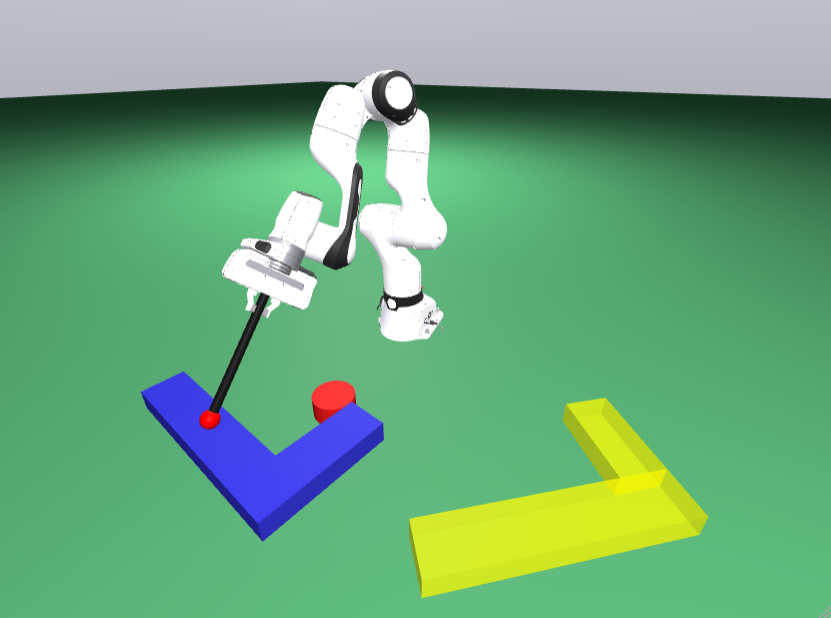}%
    \vspace{-2mm} 
    \caption{A single arm uses press-and-slide to transport an object, then leverages it as a tool to reposition a second object.}
    \label{fig:exp0}
    \vspace{-4mm} 
\end{figure}

Non-prehensile planar manipulation is a fundamental capability for robots operating in homes, warehouses, factories, and service environments. In many practical situations, a robot must move an object without forming a stable grasp: the object may be too large, too heavy, poorly shaped, cluttered by surrounding items, or simply more efficiently moved by pushing or sliding than by lifting. For instance, an object can be transported and reoriented through planar contact, then used as a tool to pull or reposition other objects (Fig.~\ref{fig:exp0}).

At the same time, planar contact manipulation is difficult to model, plan, and control. Consequently, pushing and press-and-slide behaviors are hybrid, underactuated, and friction-limited, making them difficult to plan globally and often expensive to control online \cite{pusherSlider}. These difficulties become more pronounced in multi-contact settings, where the robot can influence not only the net object wrench but also the effective center of pressure through the distribution of normal forces. Nevertheless, this additional authority also creates new manipulation opportunities, including tighter reorientation maneuvers, pivoting about a chosen contact region, and more controlled wall- or surface-assisted transport.

In this paper, we argue that many practically relevant planar contact interactions admit a simpler interpretation than a fully generic hybrid contact problem. Our key idea is to organize the planar manipulation into a small set of physically meaningful contact configurations and to associate each configuration with a reduced-order motion template. Following the idea of single-arm rear pushing being modeled as a bounded-curvature car model \cite{pushingRevisited}, we make a comprehensive analysis of manipulation tasks with single and bimanual arms.

This perspective leads to a \textit{mode-aware} pipeline: given a task, the robot selects a contact configuration, plans motion in the corresponding reduced model, recovers the required object wrench/twist, and allocates feasible forces to one or two arms under friction and unilateral constraints. For long-horizon motions, robot kinematic feasibility is incorporated into planning by optimizing robot configurations while keeping the end-effector contact pose fixed in object coordinates.

\begin{table*}[t] 
\centering
\vspace{-2mm}
\caption{Summary of Planar Contact Manipulation Modes and Kinematic Properties}
\label{tab:mode_summary}
\vspace{-2mm}
\begin{tabular}{@{}llcc@{}}
\toprule
\textbf{Manipulation Mode} & \textbf{Reduced-Order Model} & \textbf{Body-Fixed Tracking Point} & \textbf{CoP Steering} \\ \midrule
Single-Arm Rear Pushing & Dubins Bicycle & \checkmark & -- \\
Single-Arm Press-and-Slide & Unicycle & \checkmark & \checkmark \\
Parallel Bimanual Pushing & Dubins Bicycle / Differential Drive & \checkmark & -- \\
Orthogonal Bimanual Pushing & Quasi-Holonomic & -- & -- \\
Bimanual Press-and-Slide & Quasi-Holonomic & \checkmark & \checkmark \\ \bottomrule
\end{tabular}
\vspace{-4mm} 
\end{table*}

\paragraph{Contributions}
This paper makes the following contributions:
\begin{itemize}
    \item \textbf{Body-fixed tracking point for press-and-slide:} We identify a body-fixed tracking point for single-arm press-and-slide motion that is invariant to the force direction. It allows planning as a unicycle, as a car, or a forklift; spanning the whole steering regime.
    \item \textbf{Mode-aware reduced-order viewpoint:} We unify common planar contact interactions under a small set of physically meaningful configurations, each admitting a simple reduced-order motion template.
    \item \textbf{Optimization-free execution:} We present closed-form, optimization-free control strategies for motion planning. Extensive simulations demonstrate our method's planning and control effectiveness.
\end{itemize}




\section{RELATED WORK}
\label{sec:related_work}

Planar contact manipulation is commonly modeled under quasi-static assumptions that map frictional wrenches to object motion. Mason formalized the mechanics of pushing and the role of contact location and friction constraints \cite{mechanicsAndPlanning}. The limit surface then provided a compact representation of support friction and the coupled translation--rotation behavior in planar sliding \cite{planarSliding,howe1996kinematics}.

A substantial literature treats planar pushing as an underactuated, nonholonomic system. \emph{Stable Pushing} identifies stable pushing directions that keep the object fixed relative to the pusher and enables obstacle-aware planning \cite{stablePushing}. \emph{Pushing Revisited} shows that sticking pushes admit a bounded-curvature car-like abstraction and a differential-flatness view for planning and stabilization \cite{pushingRevisited}. Feedback-control formulations further emphasize hybrid contact modes and friction-limited inputs in the pusher--slider \cite{pusherSlider}, motivating optimization-based controllers such as hybrid MPC for online reasoning over mode sequences \cite{hogan2020reactive}.

Motion-cone formulations offer another compact reachability abstraction under friction \cite{motionCones}. Top-contact planar sliding has received comparatively less attention: Yi and Fazeli analyze top-contact sliding via asymmetric dual limit surfaces and derive goal-reaching planners under top-contact constraints \cite{dualLimitSurfaces}. Relatedly, pulling under fixed contact can exploit stabilizing behavior when the center of pressure trails the pulling direction; Huang \emph{et al.} derive exact bounds under pressure uncertainty for guaranteed plans \cite{exactBounds}.

Beyond analytical pushing, surveys synthesize contact-rich manipulation methods across modeling, hybrid/dynamic control, simulation, and learning \cite{pushSurvey,nonprehensileSurvey}. In clutter, push-grasping combines pushing with grasp planning \cite{pushGrasping}. Data-driven models have been integrated with MPC for accurate closed-loop pushing with limited data \cite{dataEfficient}, supported by large-scale datasets and probabilistic predictors of outcomes and variability \cite{millionWays,probabilisticPushing}. Learning has also coupled pushing and grasping via multi-task and self-supervised RL \cite{pintoGupta,zengSynergy}, and recurrent models such as Push-Net infer latent object properties from push history \cite{pushNet}. More broadly, contact-mode guided planning searches over hybrid contact sequences in 3D \cite{contactModeGuided}, dynamic in-hand sliding incorporates inertia with limit-surface models \cite{dynamicSliding}, and tool-assisted push--pull primitives enable confined manipulation \cite{toolPushPull}.

Our work builds on these foundations, but rather than specializing in single primitives, we unify single and bimanual pushing and press-and-slide maneuvers into a mode-aware reduced-order framework, enabling closed-form force allocation with optional kinematic feasibility planning for long horizons.


\section{BACKGROUND INFORMATION}
\label{sec:background}

\subsection{Planar Kinematics and the Limit Surface}
\label{subsec:limit_surface_kinematics}

We model planar motion under the quasi-static assumption, where frictional forces dominate inertial forces. Let the net planar wrench applied to the object be $W = [F_x, F_y, \tau]^\top$, and the resulting planar twist be $\nu = [v_x, v_y, \omega]^\top$. In quasi-static equilibrium, support friction perfectly opposes $W$. The boundary of frictional wrenches the contact patch can sustain is defined by the \emph{limit surface}, standardly approximated as an ellipsoid:
\begin{equation}
H(W) \;=\; W^\top A\,W \;=\; 1,
\label{eq:ls_ellipsoid}
\end{equation}

where $A$ is a diagonal shape matrix. For isotropic friction, $A$ uses shape coefficients $a$ and $b$:
\begin{equation}
A \;=\; \mathrm{diag}(a, a, b).
\label{eq:A_matrix}
\end{equation}
These coefficients represent the inverse-squared semi-axis lengths of the ellipsoid:
\begin{equation}
a \;=\; \frac{1}{(\mu N_{\mathrm{total}})^2}, \qquad b \;=\; \frac{1}{(c\,r_0\,\mu N_{\mathrm{total}})^2}.
\label{eq:ab_defs}
\end{equation}
Here, $\mu$ is the sliding friction coefficient, $N_{\mathrm{total}}$ is the total normal force, $r_0$ is the effective contact radius, and $c \in (0,1]$ is a dimensionless pressure distribution constant (typically $c \approx 0.6$).

During active sliding, the instantaneous twist aligns with the outward normal of the limit surface at the realized wrench (\emph{normality rule}):
\begin{equation}
\nu \;=\; \lambda \nabla H(W) \;=\; 2\lambda A W, \qquad \lambda > 0.
\label{eq:normality}
\end{equation}
This induces a local linear mapping between the applied wrench and the resulting motion:
\begin{equation}
v_x = \alpha F_x, \qquad v_y = \alpha F_y, \qquad \omega = \beta \tau,
\label{eq:mobility_map}
\end{equation}
where the translational ($\alpha$) and rotational ($\beta$) mobilities are:
\begin{equation}
\begin{aligned}
\alpha &\;=\; 2\lambda a \;=\; \frac{2\lambda}{(\mu N_{\mathrm{total}})^2}, \\
\beta &\;=\; 2\lambda b \;=\; \frac{2\lambda}{(c\,r_0\,\mu N_{\mathrm{total}})^2} \;=\; \frac{\alpha}{(c\,r_0)^2}.
\end{aligned}
\label{eq:alpha_beta}
\end{equation}
Thus, for a fixed contact geometry, translational and rotational mobilities are physically coupled, differing strictly by the characteristic squared length scale $(c\,r_0)^2$.

\subsection{Center of Pressure and Normal Load Scaling}
\label{subsec:cop_scaling}

The structure of the limit surface is fundamentally anchored to the \emph{Center of Pressure} (CoP). For an object with weight $N_{\mathrm{obj}}$ at $(x_{\mathrm{obj}}, y_{\mathrm{obj}})$, subjected to $k$ discrete downward robotic contact forces $N_i$ at $(x_i, y_i)$, the total normal force is:
\begin{equation}
N_{\mathrm{total}} \;=\; N_{\mathrm{obj}} + \sum_{i=1}^{k} N_i.
\label{eq:total_normal}
\end{equation}
The spatial location of the CoP, $(x_{\mathrm{CoP}}, y_{\mathrm{CoP}})$, is strictly defined by the normal-force-weighted average of these contact locations:
\begin{equation}
\begin{aligned}
x_{\mathrm{CoP}} &\;=\; \frac{x_{\mathrm{obj}} N_{\mathrm{obj}} + \sum_{i=1}^{k} x_i N_i}{N_{\mathrm{total}}}, \\
y_{\mathrm{CoP}} &\;=\; \frac{y_{\mathrm{obj}} N_{\mathrm{obj}} + \sum_{i=1}^{k} y_i N_i}{N_{\mathrm{total}}}.
\end{aligned}
\label{eq:cop_coordinates}
\end{equation}
In robotic manipulation, particularly bimanual top-contact pressing ($k=2$), both the CoP location and the applied normal load can be actively modulated. 

Assuming the patch geometry $(c, r_0)$ remains approximately constant, \eqref{eq:alpha_beta} shows both mobility constants scale inversely with the square of the total normal force:
\begin{equation}
\alpha \propto \frac{1}{N_{\mathrm{total}}^2},\qquad \beta \propto \frac{1}{N_{\mathrm{total}}^2}.
\label{eq:N2_scaling}
\end{equation}

A pure spatial shift of the CoP alters the moment arm for externally applied forces but does not fundamentally change the local mobility coefficients ($\alpha, \beta$). Crucially, when an applied action simultaneously shifts the CoP and increases the local normal force, the physical deformation of the limit surface is dominantly driven by the $1/N_{\mathrm{total}}^2$ scaling. Consequently, in dynamic top-pressing, the CoP acts primarily as a geometric control variable to redirect torque, while the total normal force dictates the object's absolute slip resistance.

\section{METHODOLOGY}
\label{sec:methodology}

Building upon the quasi-static friction mechanics established in Section~\ref{sec:background}, we abstract complex planar limit surface mappings into highly tractable reduced-order models (ROMs) (Table \ref{tab:mode_summary}). By algebraically isolating points of zero lateral slip, we map continuous frictional mechanics to classic non-holonomic wheeled-vehicle analogues. The following subsections systematically analyze five contact topologies, demonstrating the evolution from strictly under-actuated, single-contact constrained models to fully expressive, quasi-holonomic platforms via active Center of Pressure (CoP) steering.

\subsection{Case 1: Single-Arm Rear Pushing}
\label{subsec:case1_rear_push}

\begin{figure}[t]
    \centering
    \begin{minipage}[b]{0.48\columnwidth}
        \centering
        \includegraphics[width=\textwidth]{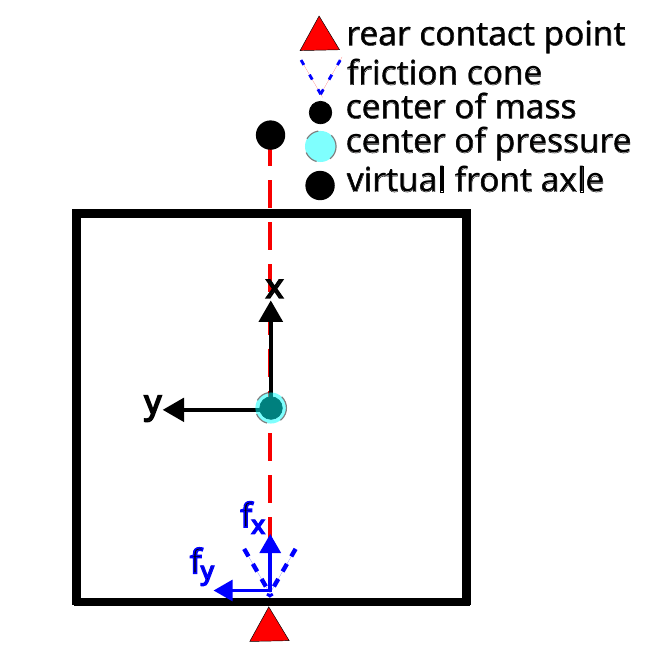}
        \vspace{-4mm}
        \centerline{(a) Symmetric contact}
        \label{fig:case1_sym}
    \end{minipage}
    \hfill
    \begin{minipage}[b]{0.48\columnwidth}
        \centering
        \includegraphics[width=\textwidth]{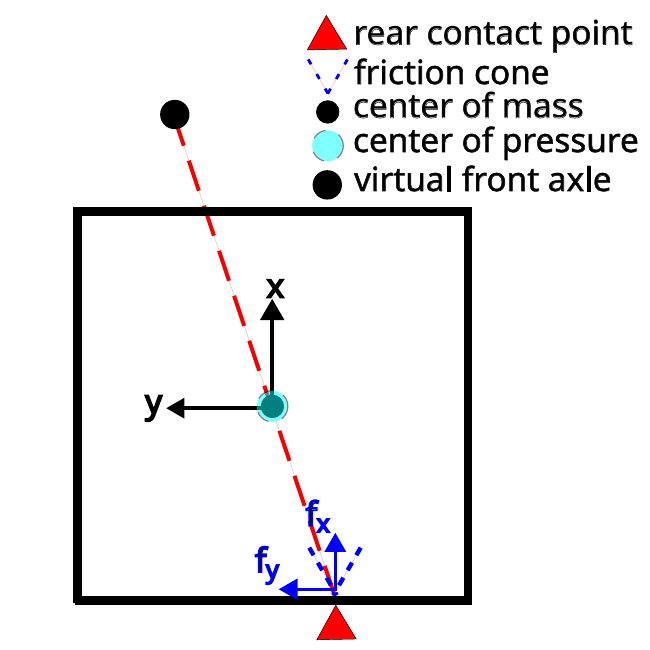}
        \vspace{-4mm}
        \centerline{(b) Unsymmetric contact}
        \label{fig:case1_asym}
    \end{minipage}\vspace{-2mm}
    \caption{Kinematic abstractions of single-arm rear pushing. (a) Symmetric pushing yields a Rear-Wheel Steering (RWS) model. (b) A lateral offset introduces a bias, distorting the steering envelope.}
    \label{fig:case1_combined}
    \vspace{-2mm}
\end{figure}

In single-arm rear pushing, the contact is located at $[-d, 0]^\top$. The applied wrench $W = [f_x, f_y, -d f_y]^\top$ bounded by $|f_y| \le \mu f_x$ generates the object's twist via the linear mobility map \eqref{eq:mobility_map}:
\begin{equation}
v_x \;=\; \alpha f_x, \qquad v_y \;=\; \alpha f_y, \qquad \omega \;=\; -\beta d f_y.
\label{eq:case1_twist}
\end{equation}
We define an effective steering angle $\delta$, bounded by the friction cone $\delta_{\max} = \pm \arctan(\mu)$:
\begin{equation}
\tan \delta \;=\; \frac{v_y}{v_x} \;=\; \frac{f_y}{f_x}.
\label{eq:steering_angle}
\end{equation}
This consequently bounds the maximum achievable path curvature $\kappa$:
\begin{equation}
\kappa \;=\; \frac{\omega}{v_x} \;=\; -\frac{\beta d}{\alpha} \tan \delta \;=\; -\frac{d}{(c r_0)^2} \tan \delta.
\label{eq:curvature_limit}
\end{equation}
Setting the lateral slip to zero ($v_y + x_{\mathrm{vfa}} \omega = 0$) yields the Virtual Front Axle (VFA):
\begin{equation}
x_{\mathrm{vfa}} \;=\; \frac{\alpha}{\beta d} \;=\; \frac{(c r_0)^2}{d}.
\label{eq:virtual_axle}
\end{equation}
The system perfectly abstracts as a Rear-Wheel Steering (RWS) Car (Fig.~\ref{fig:case1_combined}a), while an off-center contact (Fig.~\ref{fig:case1_combined}b) introduces an asymmetric normal force bias. Because RWS kinematics are non-minimum phase, autonomous trajectory tracking must regulate the VFA via a rear-wheel adapted Stanley control law to guarantee asymptotic stability without oscillation \cite{Stanley}:
\begin{equation}
\delta \;=\; -\theta_e - \arctan\left(\frac{k e_{\mathrm{vfa}}}{v_x}\right),
\label{eq:rws_stanley}
\end{equation}
where $k>0$ is a tuning gain, $\theta_e$ is heading error, and $e_{\mathrm{vfa}}$ is the cross-track error at the VFA.

\subsection{Case 2: Single-Arm Top Press-and-Slide}
\label{subsec:case2_top_press}

\begin{figure}[t]
    \centering
    \begin{minipage}[b]{0.48\columnwidth}
        \centering
        \includegraphics[width=\textwidth]{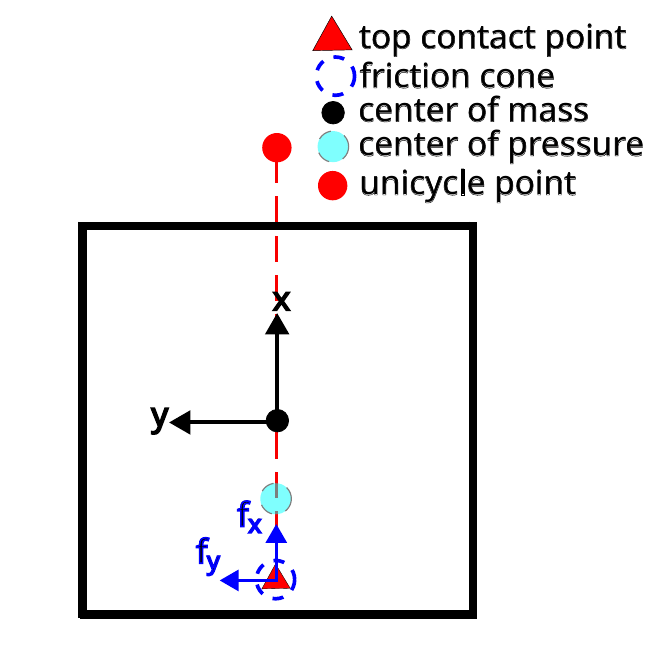}
        \vspace{-4mm}
        \centerline{(a) Symmetric contact}
        \label{fig:case2_sym}
    \end{minipage}
    \hfill
    \begin{minipage}[b]{0.48\columnwidth}
        \centering
        \includegraphics[width=\textwidth]{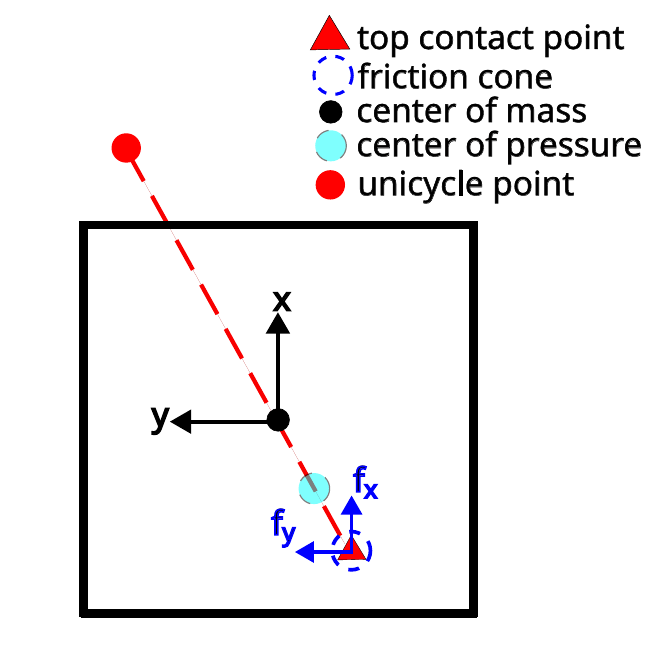}
        \vspace{-4mm}
        \centerline{(b) Unsymmetric contact}
        \label{fig:case2_asym}
    \end{minipage}\vspace{-2mm}
    \caption{Single-arm top press-and-slide. The virtual axle lies on the longitudinal axis for symmetric contacts (a) and diametrically opposite the CoM for unsymmetric contacts (b).}
    \label{fig:case2_combined}
    \vspace{-2mm}
\end{figure}

\begin{figure*}[t]
    \centering
    \includegraphics[width=0.64\textwidth]{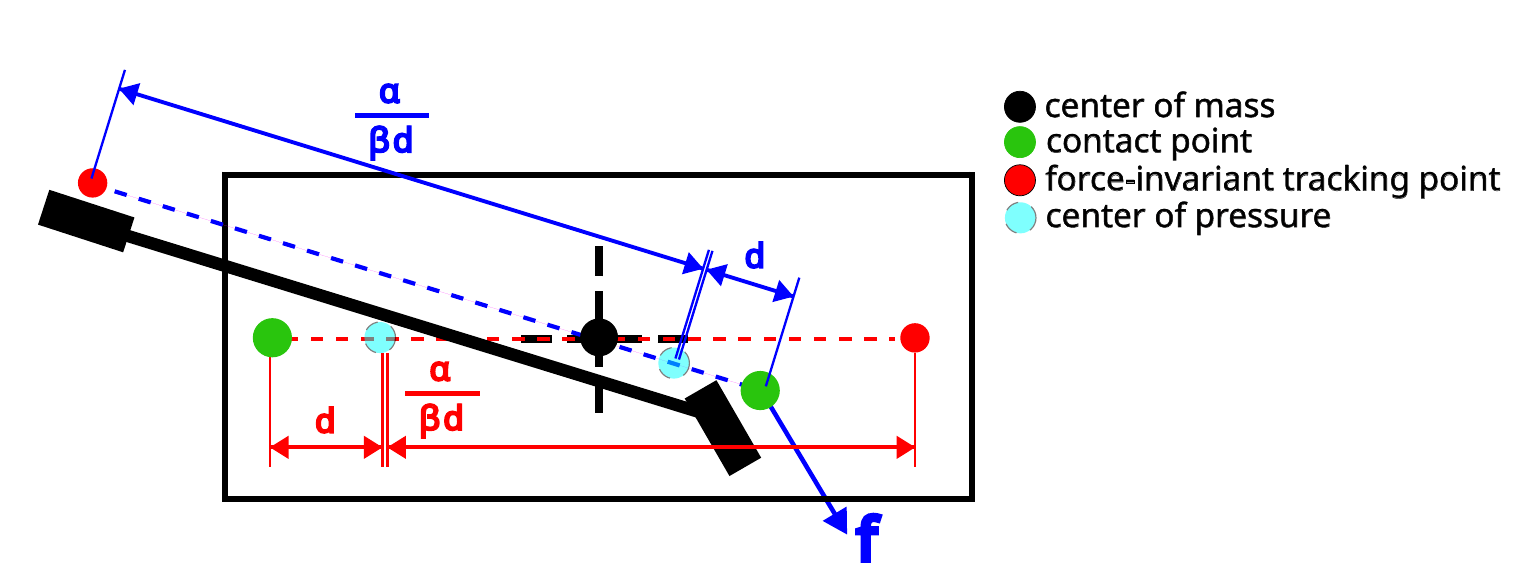}
    \vspace{-2mm}
    \caption{Top press-and-slide motion admits a planar force-invariant, body-fixed tracking point. This geometric point remains kinematically valid for all applied planar forces within the 2D friction circle; and acts like a virtual axle, allowing motion planning as a unicycle, car, or forklift.}
    \label{fig:hero_virtual_axle}
    \vspace{-4mm}
    
\end{figure*}

Elevating actuation to top-contact at $[d, 0]^\top$ (on the object longitudinal axis) allows a planar force vector $\mathbf{F}_c = [F_x, F_y]^\top$ bounded by a 2D friction circle. The resulting twist is:
\begin{equation}
v_x \;=\; \alpha F_x, \qquad v_y \;=\; \alpha F_y, \qquad \omega \;=\; \beta d F_y.
\label{eq:case2_twist}
\end{equation}
Solving for the coordinate $x_{\mathrm{va}}$ where lateral slip is zero yields:
\begin{equation}
x_{\mathrm{va}} \;=\; -\frac{v_y}{\omega} \;=\; -\frac{\alpha F_y}{\beta d F_y} \;=\; -\frac{(c r_0)^2}{d}.
\label{eq:case2_virtual_axle}
\end{equation}
Crucially, the lateral force $F_y$ cancels entirely; $x_{\mathrm{va}}$ is a static geometric constant dependent strictly on limit surface parameters and arm placement (Fig.~\ref{fig:hero_virtual_axle}). This single invariant Virtual Axle remains a valid non-holonomic stabilization center across the entire steering regime, perfectly bridging car-like translation and unicycle-like pure pivoting. The sign of $d$ strictly dictates intrinsic stability: pulling ($d > 0$) yields a naturally stable Front-Wheel Steering system, whereas pushing ($d < 0$) exhibits non-minimum phase behavior and consequently requires better control. This invariant point permits a unified $(v, \omega)$ kinematic control interface, entirely bypassing the need for piecewise mode-switching heuristics. For any contact location, the virtual axle lies on the line connecting the contact to the CoM (Fig.~\ref{fig:case2_combined}); its position is force-invariant but shifts with pressing force via CoP displacement.

\subsection{Case 3: Dual-Arm Rear Pushing}
\label{subsec:case3_dual_rear}

\begin{figure}[t]
    \centering
    \begin{minipage}[b]{0.48\columnwidth}
        \centering
        \includegraphics[width=\textwidth]{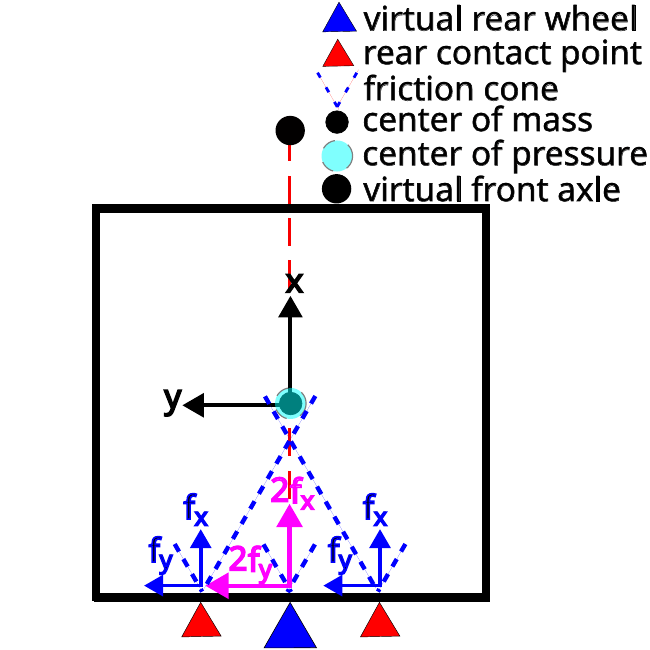}
        \vspace{-4mm}
        \centerline{(a) Equivalent bicycle}
        \label{fig:case3_bike}
    \end{minipage}
    \hfill
    \begin{minipage}[b]{0.48\columnwidth}
        \centering
        \includegraphics[width=\textwidth]{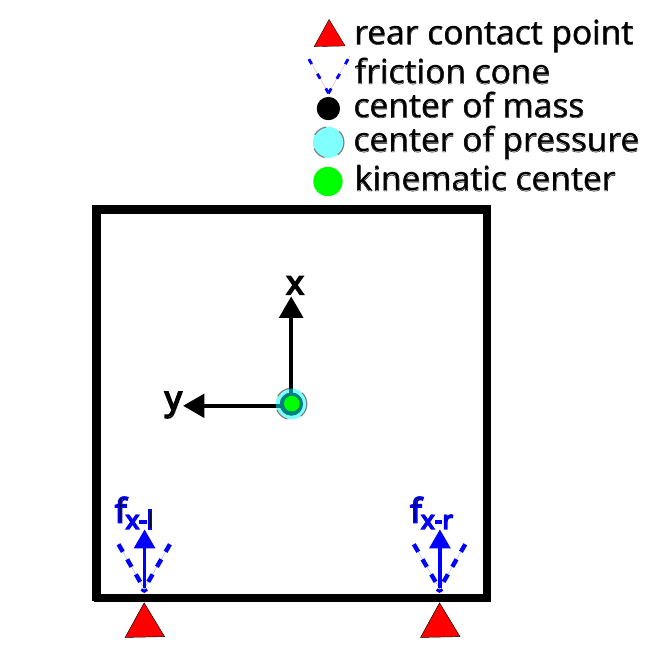}
        \vspace{-4mm}
        \centerline{(b) Differential drive}
        \label{fig:case3_diff}
    \end{minipage}\vspace{-2mm}
    \caption{Dual-arm rear pushing collapses into an Equivalent Bicycle model (a) via synchronized forces, or a Differential Drive model (b) via pure differential normal forces.}
    \label{fig:case3_combined}
    \vspace{-2mm}
\end{figure}

Introducing two rear contacts at $[-d_x, \pm w]^\top$ yields an over-actuated net wrench at the CoM:
\begin{equation}
\begin{bmatrix} F_x \\ F_y \\ \tau \end{bmatrix} \;=\; 
\begin{bmatrix} f_{xL} + f_{xR} \\ f_{yL} + f_{yR}  \\ -d_x(f_{yL} + f_{yR}) + w(f_{xR} - f_{xL})\end{bmatrix}.
\label{eq:case3_wrench}
\end{equation}
We strategically constrain this topology into two modes. In Mode 1 (Equivalent Bicycle), synchronized forces ($f_{xL} = f_{xR} = f_x$ and $f_{yL} = f_{yR} = f_y$) perfectly cancel normal force moments, yielding the twist:
\begin{equation}
v_x \;=\; \alpha (2f_x), \qquad v_y \;=\; \alpha (2f_y), \qquad \omega \;=\; -\beta d_x (2f_y).
\label{eq:case3_mode1_twist}
\end{equation}
The Virtual Front Axle collapses to the identical single-arm RWS model derived in Case 1:
\begin{equation}
x_{\mathrm{vfa}} \;=\; \frac{\alpha}{\beta d_x} \;=\; \frac{(c r_0)^2}{d_x}.
\label{eq:case3_vfa}
\end{equation}
This mode doubles the absolute wrench capacity without altering the RWS Stanley architecture (Fig.~\ref{fig:case3_combined}a). Alternatively, in Mode 2 (Differential Drive), eliminating lateral friction ($f_{yL} = f_{yR} = 0$) steers the object entirely via differential normal forces:
\begin{equation}
v_x \;=\; \alpha (f_{xL} + f_{xR}), \qquad v_y \;=\; 0, \qquad \omega \;=\; \beta w (f_{xR} - f_{xL}).
\label{eq:case3_mode2_twist}
\end{equation}
Because lateral velocity is permanently zero, the Instantaneous Center of Rotation (ICR) is constrained to $x_{\mathrm{icr}} = 0$. This perfectly mimics a classical differential drive robot (Fig.~\ref{fig:case3_combined}b), guaranteeing maneuverability in low-friction environments without risking lateral slip boundary failures.

\subsection{Case 4: Orthogonal Bimanual Pushing}
\label{subsec:case4_orthogonal}

\begin{figure}[t]
    \centering
    \begin{minipage}[b]{0.48\columnwidth}
        \centering
        \includegraphics[width=\textwidth]{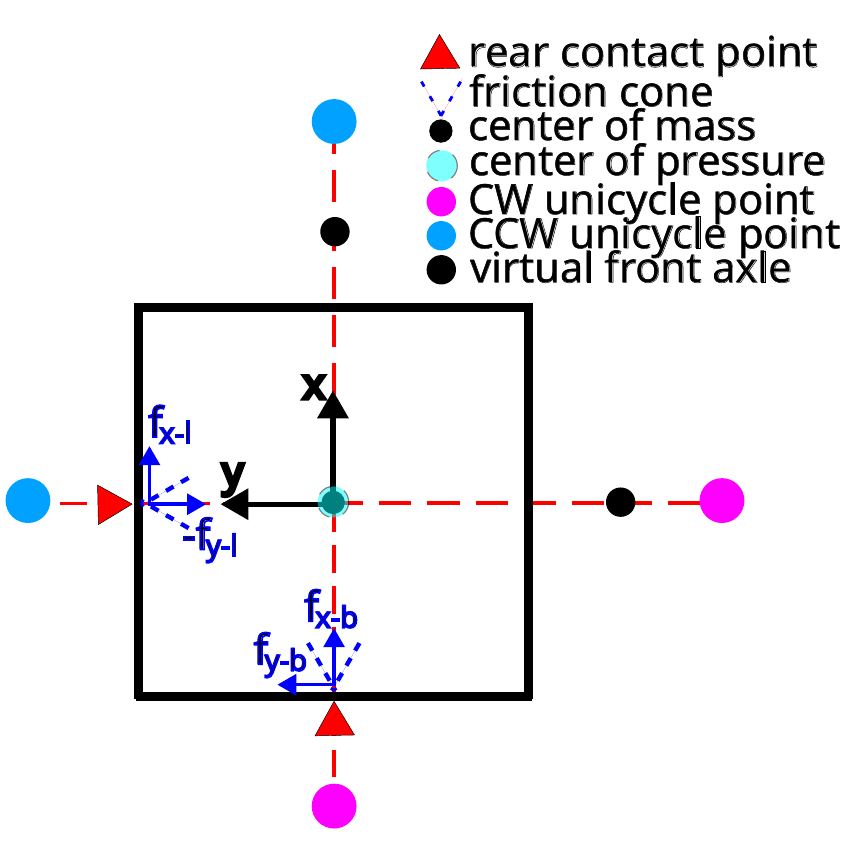}
        \vspace{-4mm}
        \centerline{(a) Extreme unicycle points}
        \label{fig:case4_points}
    \end{minipage}
    \hfill
    \begin{minipage}[b]{0.48\columnwidth}
        \centering
        \includegraphics[width=\textwidth]{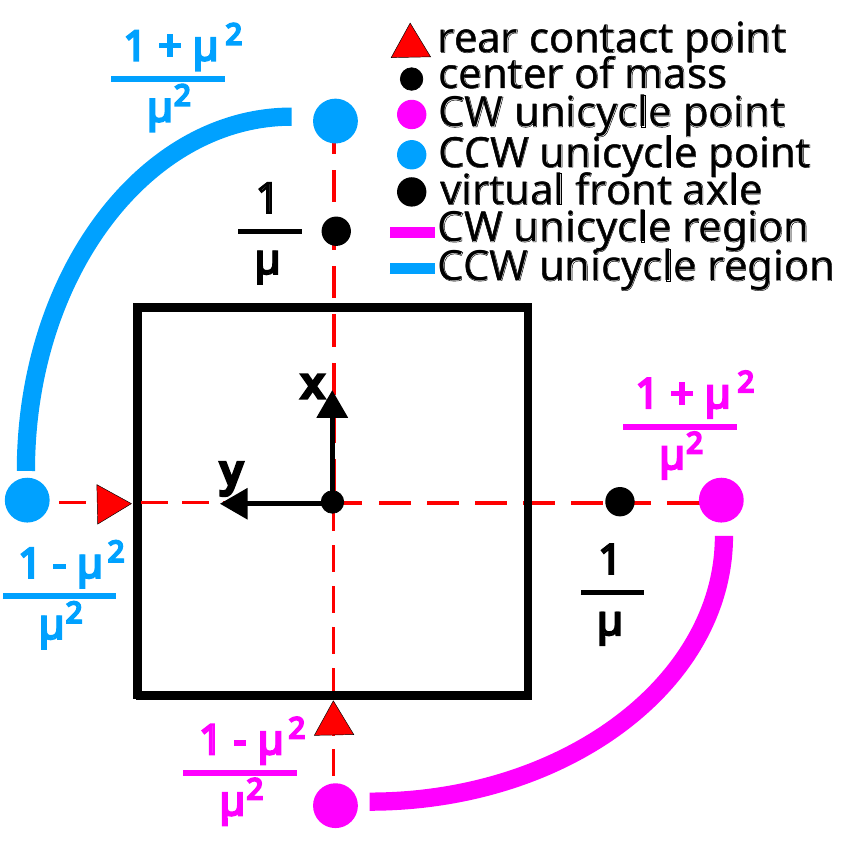}
        \vspace{-4mm}
        \centerline{(b) Unicycle regions}
        \label{fig:case4_regions}
    \end{minipage}\vspace{-2mm}
    \caption{Orthogonal bimanual pushing generates bifurcated unicycle regions depending on rotation direction, causing the ICR to teleport during motion transitions.}
    \label{fig:case4_combined}
    \vspace{-2mm}
\end{figure}

For orthogonal contacts (bottom at $x = -d$, left at $y = d$), the applied wrench is the superposition of the contact forces:
\begin{equation}
\begin{bmatrix} F_x \\ F_y \\ \tau \end{bmatrix} \;=\; 
\begin{bmatrix} f_{xB} + f_{xL} \\ f_{yB} - f_{yL} \\ -d(f_{yB} + f_{xL}) \end{bmatrix}.
\label{eq:case4_wrench}
\end{equation}
Because rotational authority is strictly generated by friction, achievable unicycle points bifurcate into two discrete, spatially isolated geometric quadrants depending on the sign of $\omega$ (CW vs. CCW) (Fig.~\ref{fig:case4_combined}). To physically illustrate this, consider pure translation in the $x$-direction: the bottom arm applies a normal force $F$, which can generate a maximum coupled lateral friction of $\pm \mu F$. The left arm must provide a balancing normal force of $\mu F$, which in turn bounds its own friction capability to $\pm \mu^2 F$. Depending on the sign of these saturated friction forces, this coupling generates two extreme unicycle points (one CW and one CCW). A symmetric analysis for pure $y$-translation yields two additional extreme points. The achievable unicycle regions span the areas strictly between these extremes. Defining the baseline kinematic length $K = \alpha/(\beta d)$, these regional boundaries fall exactly at distances of $d_{\min} = \frac{1-\mu^2}{\mu^2} K$ and $d_{\max} = \frac{1+\mu^2}{\mu^2} K$. Consequently, these regions are completely disjoint from one another and fail to encompass the standard single-arm bicycle tracking point located at $\frac{1}{\mu}K$. Transitioning between rotation directions requires the tracking point to pass through infinity, rendering classical non-holonomic tracking mathematically ill-posed. Instead, we abandon unicycle tracking and control the CoM directly via a desired stabilizing quasi-holonomic wrench $\mathbf{w}_{\mathrm{des}} = [F_x^*, F_y^*, \tau^*]^\top$. We map this to the four actuator commands using a direct $\mathcal{O}(1)$ algebraic allocator by symmetrically distributing the rotational friction:
\begin{equation}
f_{yB} \;=\; f_{xL} \;=\; -\frac{\tau^*}{2d}.
\label{eq:friction_alloc}
\end{equation}
Substituting \eqref{eq:friction_alloc} back into \eqref{eq:case4_wrench} allows us to solve for the required normal forces using a strictly positive internal bias $f_{\mathrm{bias}} \ge 0$:
\begin{equation}
\begin{aligned}
f_{xB} &\;=\; F_x^* + \frac{\tau^*}{2d} + f_{\mathrm{bias}}, \\
f_{yL} &\;=\; -F_y^* - \frac{\tau^*}{2d} + f_{\mathrm{bias}}.
\end{aligned}
\label{eq:normal_alloc}
\end{equation}
This algebraic mapping natively decouples translation and rotation, completely circumventing tracking-point singularities.

\subsection{Case 5: Dual-Arm Top Press-and-Slide}
\label{subsec:case5_top_press}

\begin{figure}[t]
    \centering
    \begin{minipage}[b]{0.48\columnwidth}
        \centering
        \includegraphics[width=\textwidth]{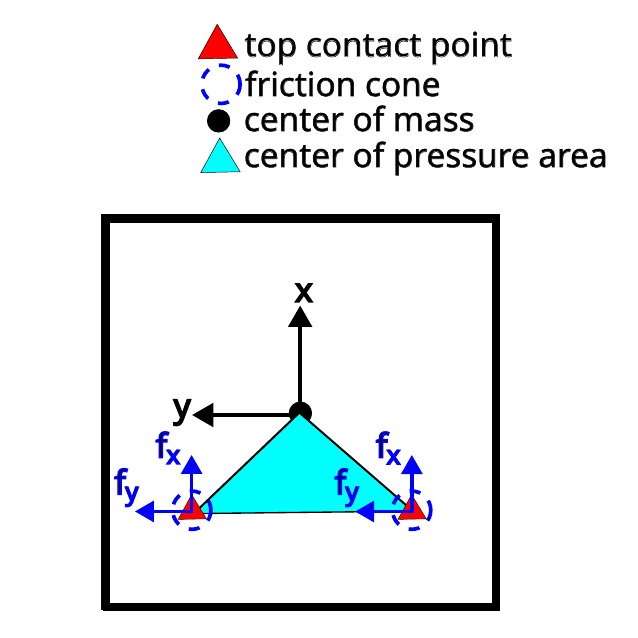}
        \vspace{-4mm}
        \centerline{(a) CoP triangle}
        \label{fig:case5_cop}
    \end{minipage}
    \hfill
    \begin{minipage}[b]{0.48\columnwidth}
        \centering
        \includegraphics[width=\textwidth]{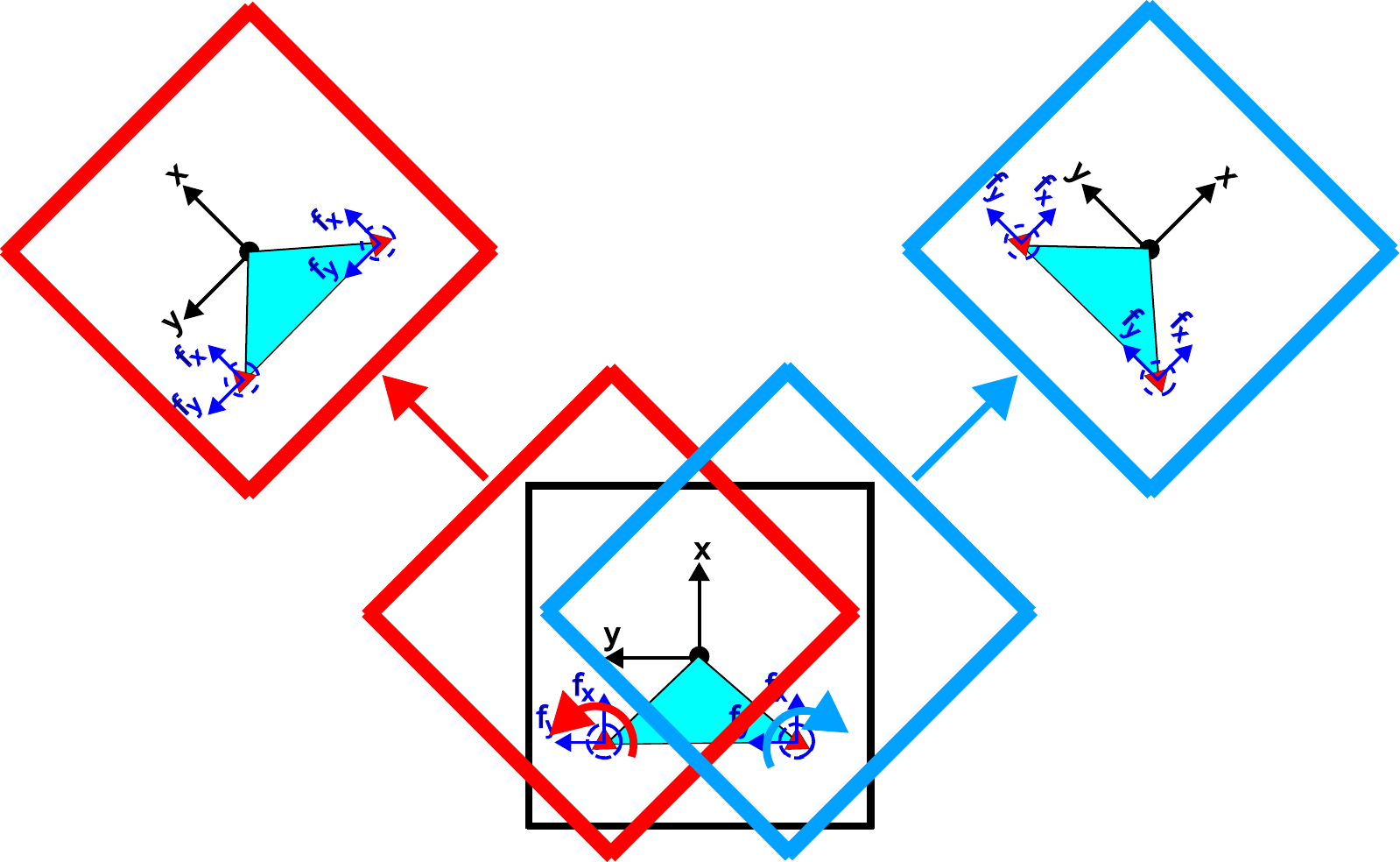}
        \vspace{-4mm}
        \centerline{(b) Pivot and move}
        \label{fig:case5_pivot}
    \end{minipage}\vspace{-2mm}
    \caption{Dual-arm top press-and-slide. The CoP is bounded within the triangular convex hull (a). Shifting the normal load allows extreme quasi-holonomic maneuvers like single-contact pivoting (b).}
    \label{fig:case5_combined}
    \vspace{-2mm}
\end{figure}

In dual-top-contact topology at $\mathbf{p}_L$ and $\mathbf{p}_R$, the spatial location of the CoP is strictly bounded within the triangular convex hull formed by the origin and the two robotic contacts (Fig.~\ref{fig:case5_combined}a). By dynamically shifting the normal load distribution between the arms, the controller actively steers the CoP, granting extreme maneuverability—such as maximizing $N_L$ and dropping $N_R \to 0$ to execute a pure unicycle pivot exactly around $\mathbf{p}_L$ (Fig.~\ref{fig:case5_combined}b). The net CoM wrench tracks a fully actuated virtual command:
\begin{equation}
\begin{bmatrix} F_x^* \\ F_y^* \\ \tau^* \end{bmatrix} \;=\; 
\begin{bmatrix} f_{xL} + f_{xR} \\ f_{yL} + f_{yR} \\ (x_L f_{yL} - y_L f_{xL}) + (x_R f_{yR} - y_R f_{xR}) \end{bmatrix}.
\label{eq:case5_wrench}
\end{equation}
To exploit this quasi-holonomic capability without optimization solvers, we resolve the redundant degree of freedom by algebraically minimizing antagonistic internal shear forces between the contacts. The required normal forces $f_{zL}, f_{zR}$ are dynamically allocated to satisfy local Coulomb friction constraints ($\mu f_{zi} \ge \|\mathbf{f}_i\|$) while deliberately placing the CoP to maximize task mobility, perfectly decoupling path planning from frictional contact mechanics.

\subsection{Mode-Aware Contact manipulation with Reduced-Order models (MACRO)}
\label{sec:framework}

The kinematic abstractions derived in Section~\ref{sec:methodology} establish a representative library of planar manipulation modalities, ranging from highly constrained under-actuated steering to fully quasi-holonomic maneuvering. These five topologies are not exhaustive but illustrate how varying contact arrangements can be systematically abstracted into reduced-order models via limit surface mechanics. 

To deploy this continuous spectrum of models on physical robotic hardware, we introduce the Mode-Aware Contact manipulation with Reduced-Order models (MACRO) framework.

\paragraph{Mode Selection and Path Planning}
Given a specific manipulation task, object properties, and environmental constraints, the first step is to choose the required contact mode. Mode selection can be optimized (e.g., minimizing path length) or dictated by feasibility, as some tasks can only be realized by the holonomic control (Fig. \ref{fig:exp4_pressfit}). The system then adopts the corresponding reduced-order model. Depending on the mode, this involves either anchoring a tracking controller to a statically valid non-holonomic point (e.g., a Virtual Axle) or abandoning non-holonomic tracking entirely in favor of Center of Mass (CoM) trajectory generation.

For long-horizon navigation, the planner computes a path incorporating both the selected model's mobility constraints and the robot kinematics. Because the end-effectors maintain fixed contact in the object's body frame, the object trajectory dictates the required arm
motions. We formulate path generation as a trajectory optimization over the manipulator joint states $\mathbf{q}$ to prevent singularities and joint-limit violations~\cite{OpAC}.

\paragraph{Wrench Allocation and Execution}
Once a reference trajectory is generated, the controller computes the required target wrench $\mathbf{W}_d$ at the designated tracking point. A critical component of the MACRO framework is the $\mathcal{O}(1)$ algebraic allocator, which resolves $\mathbf{W}_d$ into a set of commanded end-effector forces $\mathbf{F}_{\mathrm{cmd}}$, guaranteeing strictly positive normal forces and friction-cone-bounded tangential forces.

To execute these actions on physical manipulators, we employ a hybrid force-impedance control architecture. Because purely feedforward force execution is susceptible to contact loss or excessive pressure under rigid-body assumptions, the commanded normal force $f_{n,\mathrm{cmd}}$ is actively regulated using a Proportional-Integral (PI) controller based on the measured contact force $f_{n,\mathrm{meas}}$:
\begin{equation}
f_{n,\mathrm{reg}} \;=\; f_{n,\mathrm{cmd}} + K_p e_n + K_i \int e_n \, dt,
\label{eq:pi_force}
\end{equation}
where $e_n = f_{n,\mathrm{cmd}} - f_{n,\mathrm{meas}}$. The regulated Cartesian force vector $\tilde{\mathbf{F}}_{\mathrm{cmd}}$ is then projected into joint space via the Jacobian transpose, acting as the primary task wrench.

To stabilize the robot's posture while applying this wrench, the force feedforward is superimposed onto an impedance controller. The choice of impedance depends strictly on the task horizon: for local, short-horizon maneuvers, we utilize a Cartesian Impedance Controller to softly maintain the end-effector's orientation and spatial alignment with the object's body frame. Conversely, for long-horizon motions requiring complex reachability, we utilize a Joint Impedance Controller to track the kinematically feasible joint path $\mathbf{q}(t)$ generated by the planner. The final commanded joint torque is:
\begin{equation}
\boldsymbol{\tau}_{\mathrm{cmd}} \;=\; \boldsymbol{\tau}_{\mathrm{imp}} + \mathbf{J}(\mathbf{q})^\top \tilde{\mathbf{F}}_{\mathrm{cmd}} + \boldsymbol{\tau}_{\mathrm{comp}},
\label{eq:final_torque}
\end{equation}
where $\boldsymbol{\tau}_{\mathrm{imp}}$ represents the selected impedance tracking law and $\boldsymbol{\tau}_{\mathrm{comp}}$ compensates for gravity and Coriolis dynamics. This architecture robustly bridges the gap between abstract reduced-order mechanics and stable physical execution, summarizing the complete MACRO framework (Algorithm~\ref{alg:macro_b}).


\begin{algorithm}[t]
\caption{MACRO: Mode-Aware Contact manipulation with Reduced-Order models}
\label{alg:macro_b}
\begin{algorithmic}[0]
\Require Task target $\mathbf{x}^*$, object model $\mathcal{O}$, mode library $\mathcal{M}$
\Ensure Joint torques $\boldsymbol{\tau}_{\mathrm{cmd}}(t)$
\State Select contact mode $m \in \mathcal{M}$ based on task and geometry
\State Assign ROM and tracking point (VFA, VA, CoM, or CoP)
\If{mode admits CoP steering}
    \State Select CoP strategy (pivot point, balanced, etc.)
\EndIf
\State Generate reference trajectory $\mathbf{x}_{\mathrm{ref}}(t)$ in ROM state space
\If{long-horizon motion}
    \State Optimize robot joint path $\mathbf{q}(t)$ for kinematic feasibility
\EndIf
\While{$\|\mathbf{x}(t) - \mathbf{x}^*\| > \epsilon$}
    \State Compute tracking error $\mathbf{e}(t)$ at the assigned tracking point
    \State Compute desired object wrench $\mathbf{W}_d(t)$ from ROM controller
    \State Allocate $\mathbf{W}_d \to \mathbf{F}_{\mathrm{cmd}}$ s.t.\ $|\mathbf{f}_i| \le \mu N_i,\; N_i > 0$ \hfill $\triangleright\;\mathcal{O}(1)$
    \State Regulate normal forces via PI to obtain $\tilde{\mathbf{F}}_{\mathrm{cmd}}$
    \State Execute $\boldsymbol{\tau}_{\mathrm{cmd}} = \boldsymbol{\tau}_{\mathrm{imp}} + \mathbf{J}(\mathbf{q})^\top \tilde{\mathbf{F}}_{\mathrm{cmd}} + \boldsymbol{\tau}_{\mathrm{comp}}$
\EndWhile
\end{algorithmic}
\end{algorithm}

\section{EXPERIMENTS AND RESULTS}
\label{sec:experiments}

To validate the proposed control framework across different manipulation modes, we conducted comprehensive single and bimanual arm experiments in Drake, using its hydroelastic contact model to accurately compute contact wrenches.

\subsection{Single-Arm Planar Pushing}
\begin{figure}[htb]
    \centering
    \includegraphics[width=0.48\columnwidth]{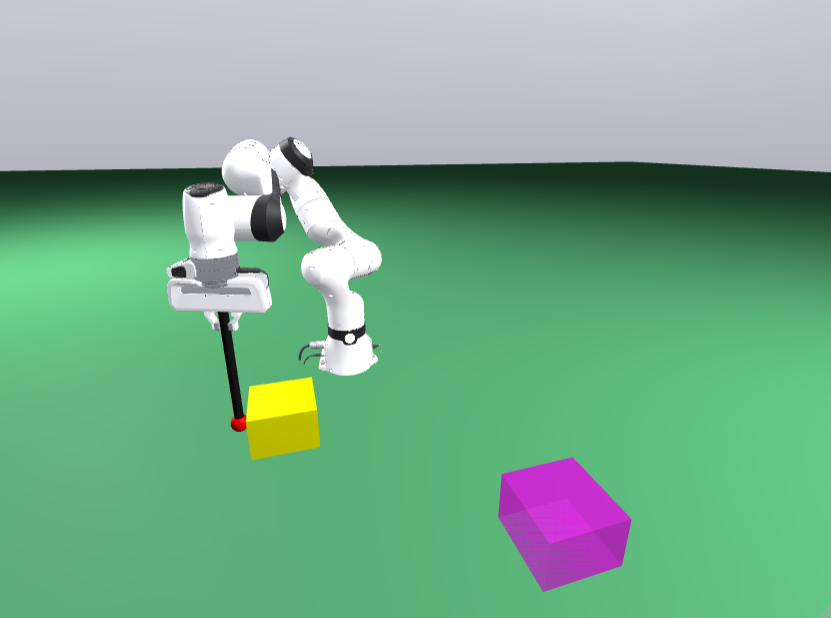}\hfill
    \includegraphics[width=0.48\columnwidth]{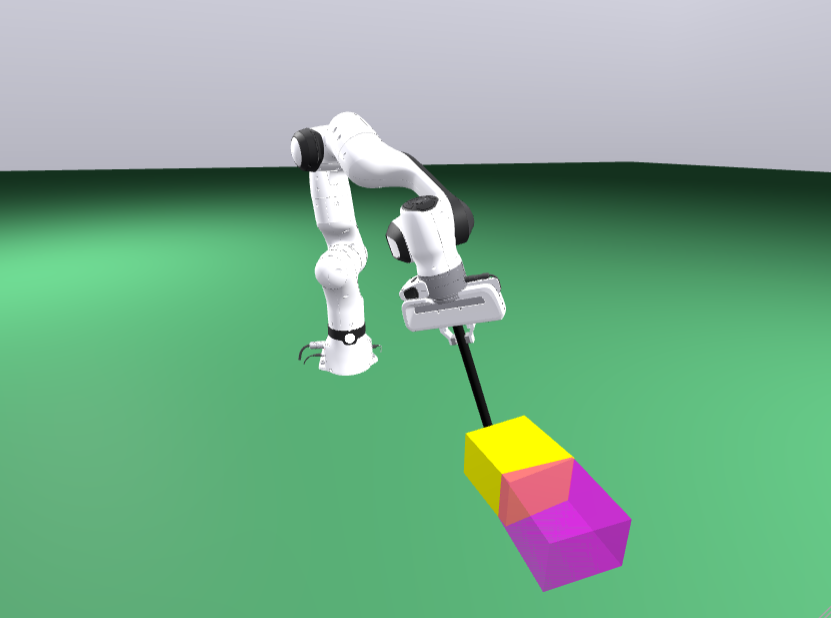}\\[1ex]
    \includegraphics[width=0.48\columnwidth]{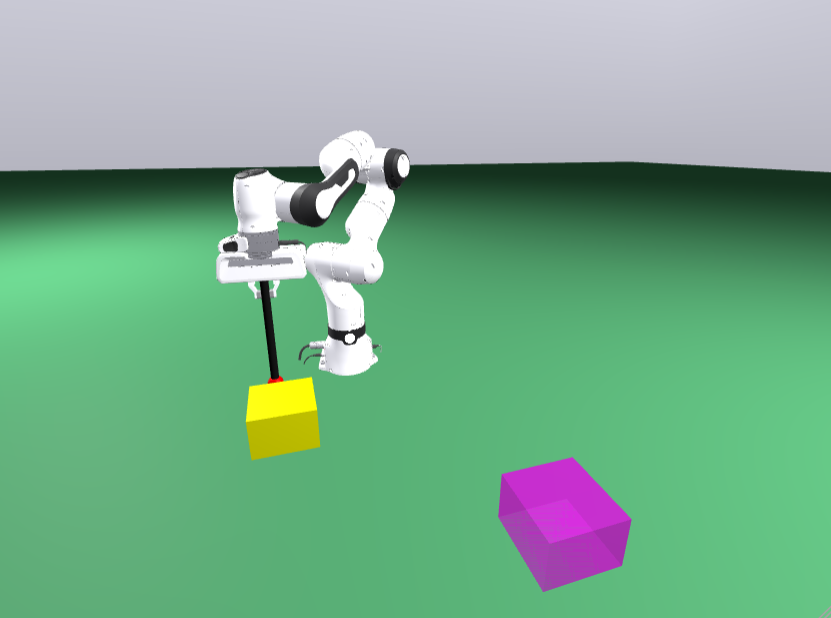}\hfill
    \includegraphics[width=0.48\columnwidth]{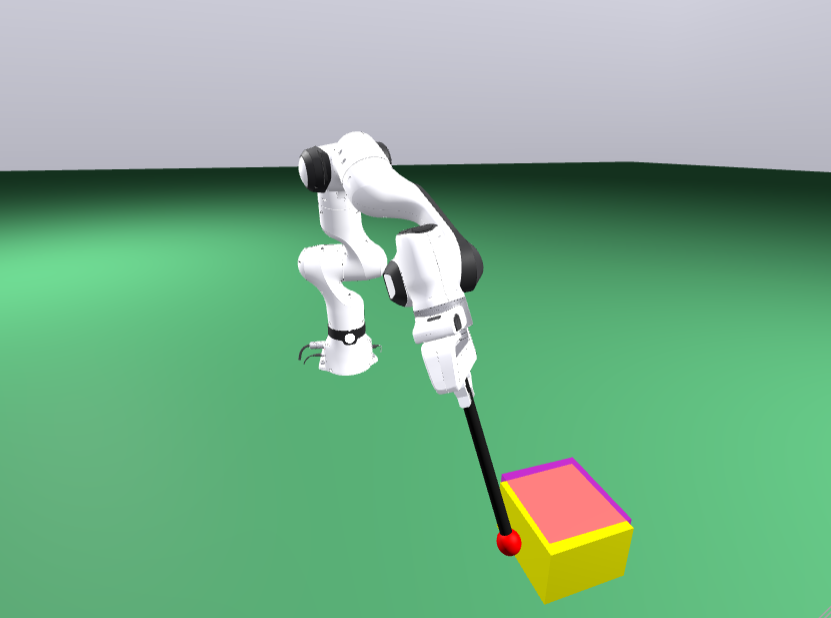}%
    \vspace{-2mm}
    \caption{Single-arm Planar Pushing. The same box is directed to the same target, using different contact points (Top: Left face, Bottom: Rear face).}
    \label{fig:exp1_push}
    \vspace{-2mm}
\end{figure}

\begin{figure}[htb]
    \centering
    \includegraphics[width=\columnwidth]{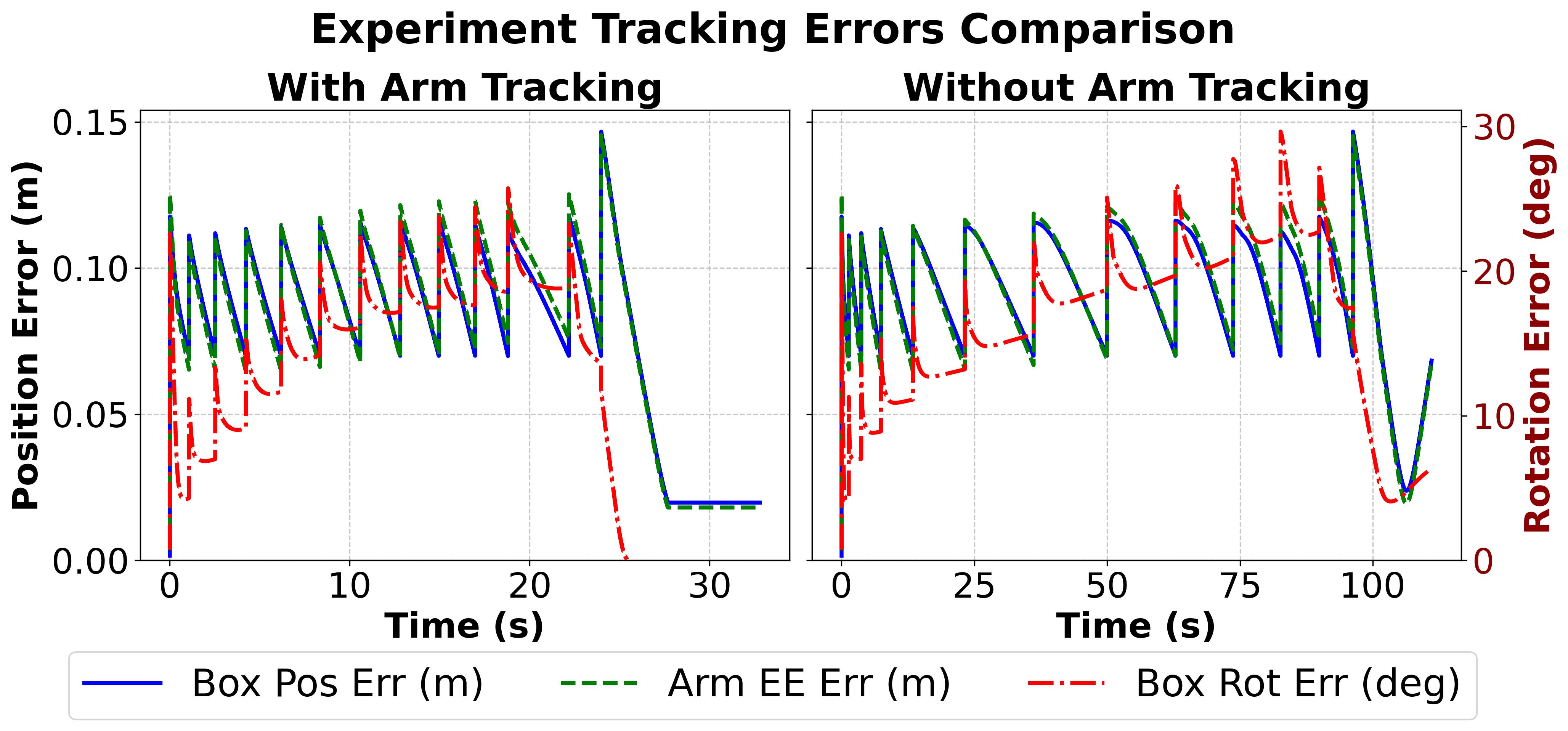}
    \vspace{-8mm}
    \caption{Comparison of tracking errors during single-arm planar pushing. Having an arm tracking term, the path is tracked faster and with less error.}
    \label{fig:exp1_rms_comparison}
    \vspace{-4mm}
\end{figure}

As shown in Fig.~\ref{fig:exp1_push}, a Franka Panda arm is tasked with pushing a 15~cm cubic box to a target pose. Because the manipulator's posture is critical in this long-horizon pushing task, Kinematic Trajectory Optimization (KTO) is used to generate a joint-space path, treating the arm and object as temporarily attached \cite{OpAC}. Pushing from two different initial sides, the controller successfully converges the box to within 1.5~cm and $2^\circ$ of the target. We observed that superimposing a low-gain proportional joint-tracking term over the primary pushing force improves the manipulator's posture against frictional disturbances without disrupting the task (Fig. \ref{fig:exp1_rms_comparison}). To quantitatively evaluate this, we executed the same trajectory with and without the joint-tracking term. When employing the joint-tracking term, the motion completed in approximately 30 seconds with root-mean-square (RMS) errors of 0.085~m for box position, 0.086~m for arm end-effector position, and $14^\circ$ for box orientation. Conversely, relying solely on the pure pushing force extended the execution time to over 100 seconds, with the respective RMS errors increasing to 0.095~m, 0.096~m, and $17.27^\circ$. This demonstrates that a moderate arm-tracking term substantially improves both execution speed and overall trajectory adherence.

\subsection{Press-and-Slide}
\begin{figure}[htb]
    \centering
    \includegraphics[width=0.48\columnwidth]{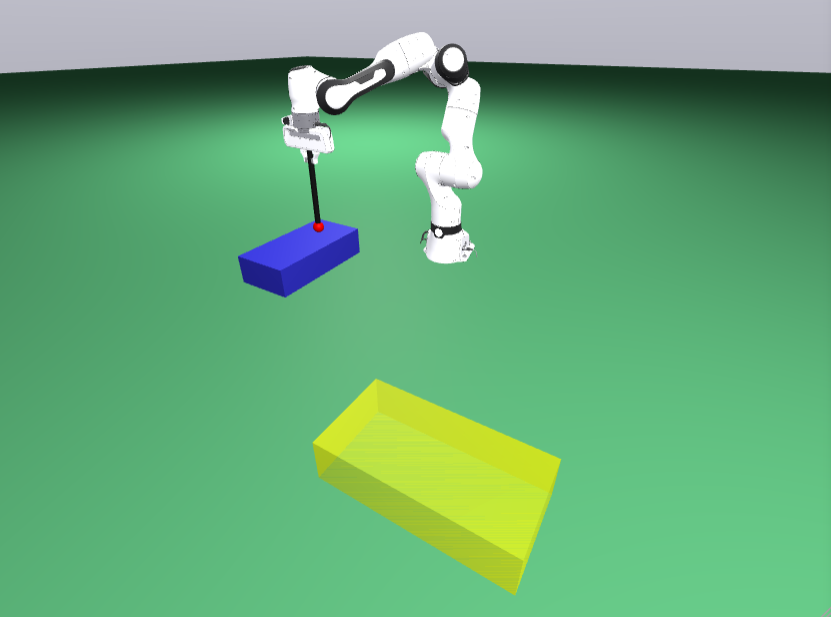}\hfill
    \includegraphics[width=0.48\columnwidth]{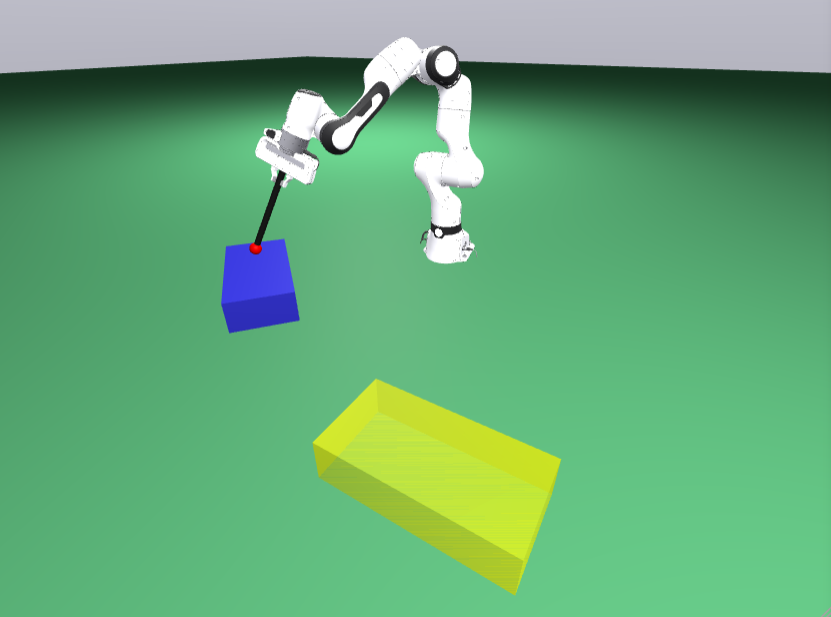}\\[1ex]
    \includegraphics[width=0.48\columnwidth]{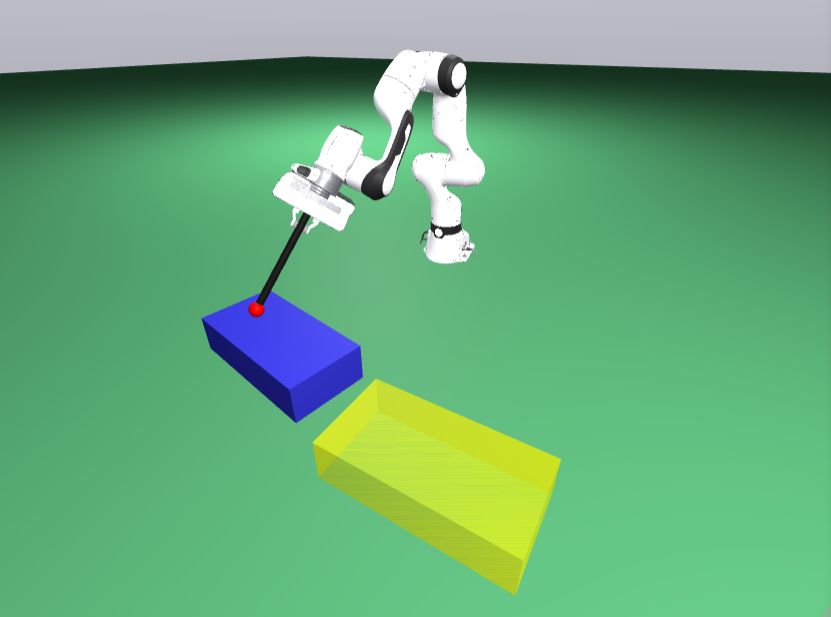}\hfill
    \includegraphics[width=0.48\columnwidth]{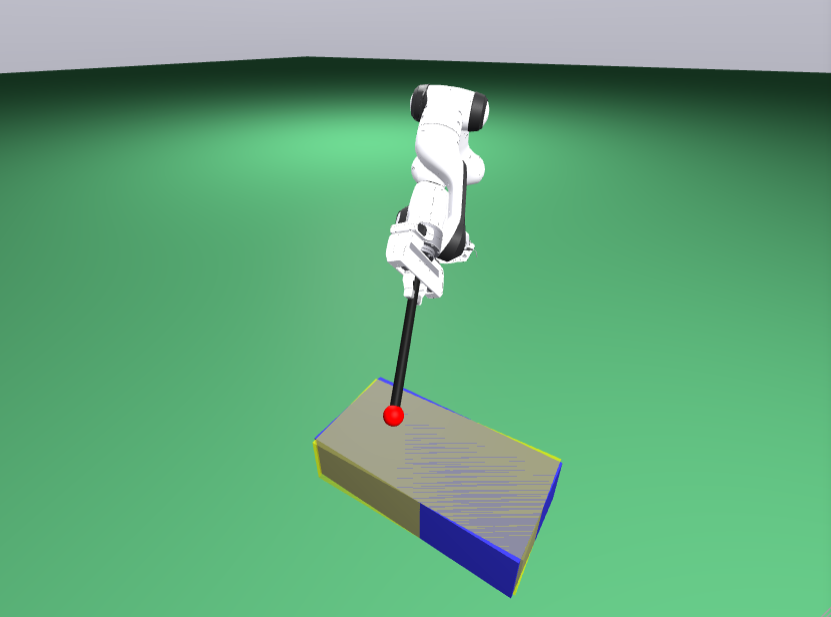}%
    \vspace{-2mm}
    \caption{Top press-and-slide motion. The box is controlled as a unicycle object, achieving desired position and orientation with a virtual target ahead of the real target.}
    \label{fig:exp2_unicycle}
    \vspace{-4mm}
\end{figure}

\begin{figure}[htb]
    \centering
    \includegraphics[width=0.8\columnwidth]{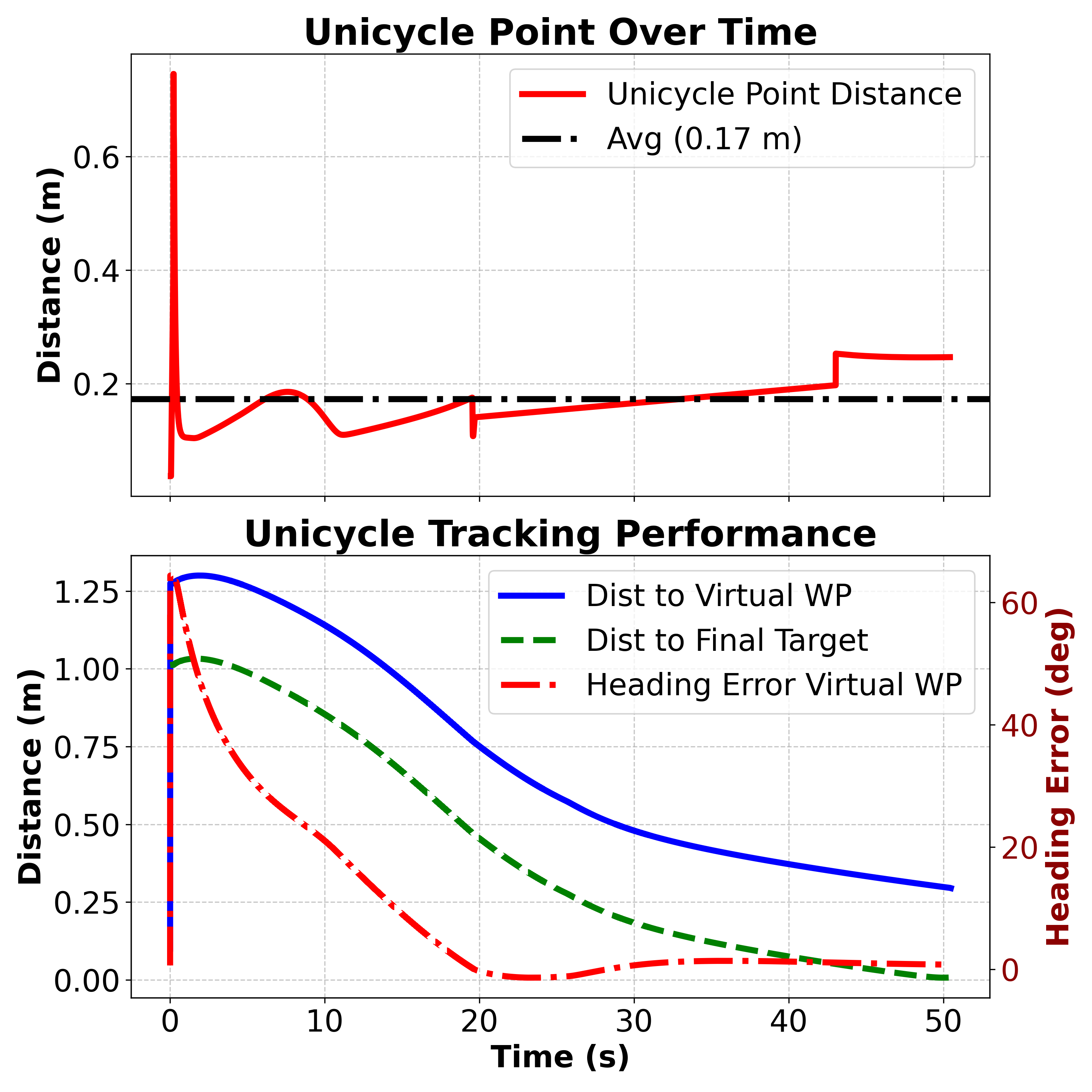}
    \vspace{-5mm}
    \caption{Unicycle point distance over time and tracking performance. Employing a virtual target, the desired one can be reached.}
    \label{fig:exp2_unicycle_plot}
    \vspace{-4mm}
\end{figure}

In this experiment, a Panda arm manipulates a $60 \times 20 \times 10$~cm rectangular box via top-pressing (Fig.~\ref{fig:exp2_unicycle}). As stated by Brockett's Theorem, a unicycle cannot be asymptotically stabilized around a specific position and orientation simultaneously using continuous feedback, due to its nonholonomic kinematics. To circumvent this limitation, the controller tracks a virtual look-ahead target projected beyond the desired pose. By applying a constant longitudinal driving force combined with a proportional lateral alignment force, the system robustly drives the object over the physical target, terminating with steady-state errors below 0.5~cm and $1^\circ$ (Fig.~\ref{fig:exp2_unicycle_plot}). 

Furthermore, we analyze the spatial stability of the force-invariant tracking point (the unicycle point), governed by the geometric ratio $\alpha/\beta = (c r_0)^2$. For the 0.3~kg box subjected to an 8~N downward arm force at a 0.2~m offset, the effective lever arm becomes $d = 0.054$~m. Computing the unicycle point under a uniform contact assumption ($c = 0.6$, $r_0^2 = 0.1$~m$^2$) yields a theoretical mobility ratio of $\alpha/\beta = 0.036$~m$^2$. Using the distance formula $\alpha/(\beta d)$, this projects the unicycle point at 0.67~m from the CoP (0.82~m from the CoM). However, the empirical unicycle point observed in simulation remains stable at approximately 0.2~m from the CoM (0.346~m from the CoP). Comparing these two results, we see that the heavy localized arm force effectively shrinks the contact patch, dropping the pressure distribution constant $c$ from its nominal 0.6 down to 0.43. This demonstrates that the virtual axle abstraction remains valid, and the press-and-slide motion can be modeled as a unicycle.

\subsection{Bimanual Pivot and Move}
\begin{figure}[htb]
    \centering
    \includegraphics[width=0.48\columnwidth]{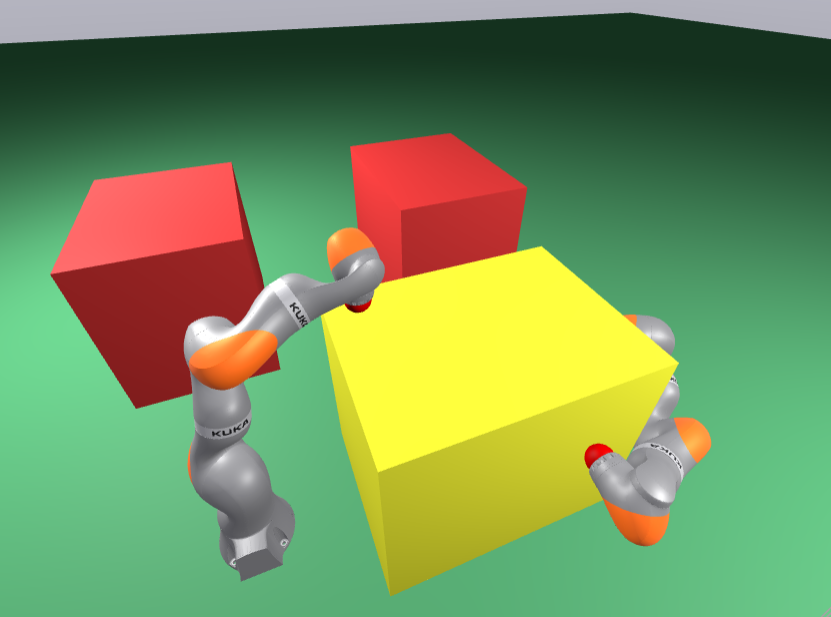}\hfill
    \includegraphics[width=0.48\columnwidth]{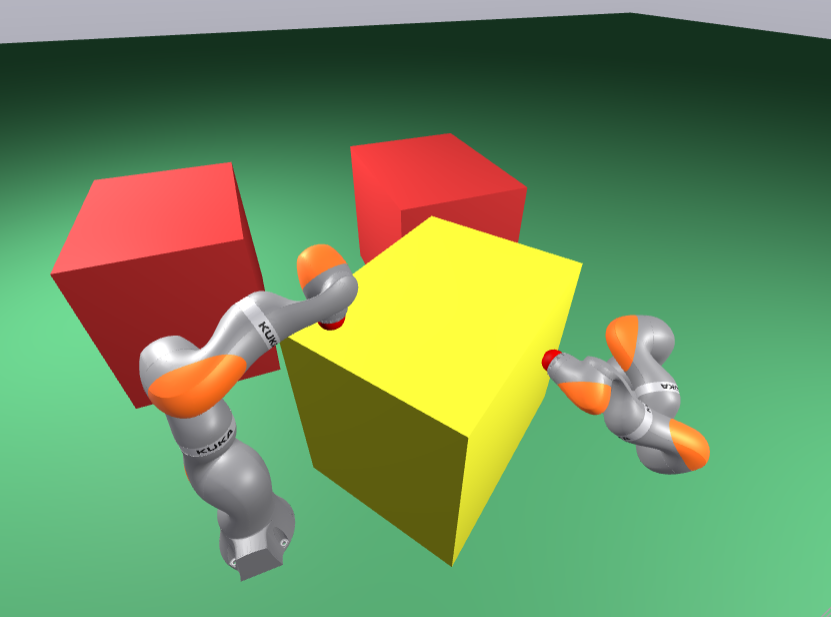}\\[1ex]
    \includegraphics[width=0.48\columnwidth]{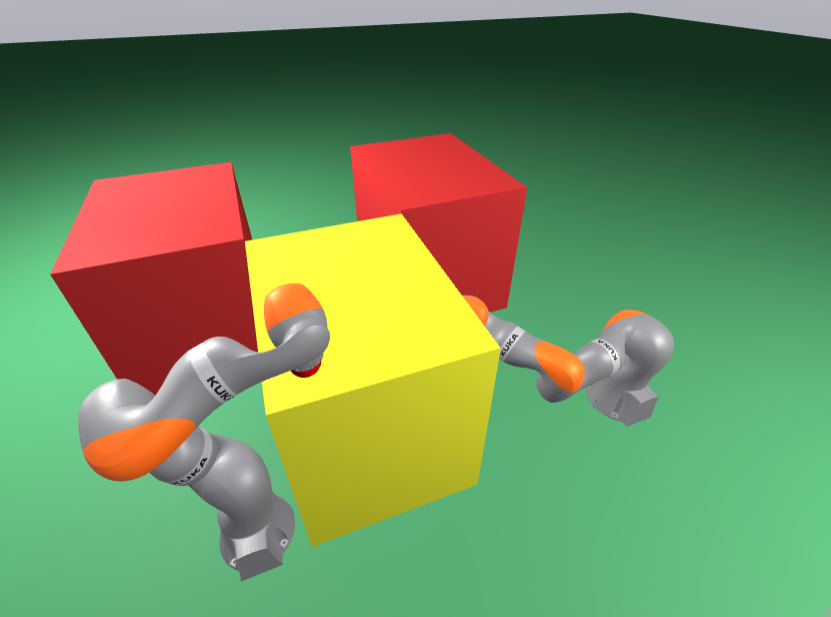}\hfill
    \includegraphics[width=0.48\columnwidth]{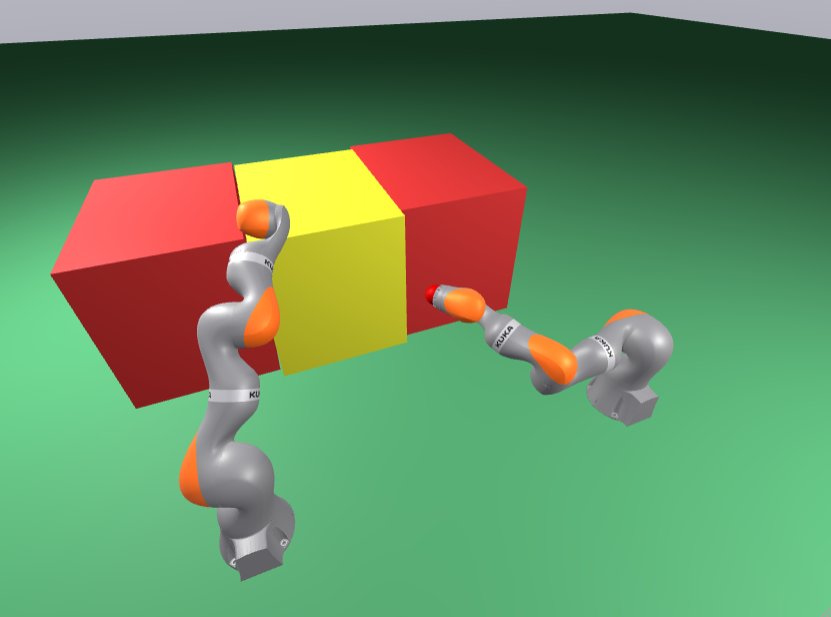}%
    \vspace{-2mm}
    \caption{Bimanual Pivot and Move. Left arm presses the box to the ground, bringing the center of pressure close to itself. After the right arm rotates the box, the left arm moves it forward to its slot.}
    \label{fig:exp3_pivot}
    \vspace{-2mm}
\end{figure}

Two KUKA iiwa7 manipulators (spaced 1.5~m apart) maneuver a large $60 \times 80 \times 70$~cm box into a tightly constrained slot between the boxes of the same size (Fig. \ref{fig:exp3_pivot}). To execute a tight turning radius, the left arm applies a dominant downward normal force, actively shifting the Center of Pressure (CoP) toward itself. Then the right arm pushes the object, pivoting it $90^\circ$ around the dynamically shifted CoP. Finally, the left arm provides a pure longitudinal push to slide the object into the slot while the right arm maintains stabilizing contact.

\subsection{Bimanual Press-Fit}
\begin{figure}[htb]
    \centering
    \includegraphics[width=0.48\columnwidth]{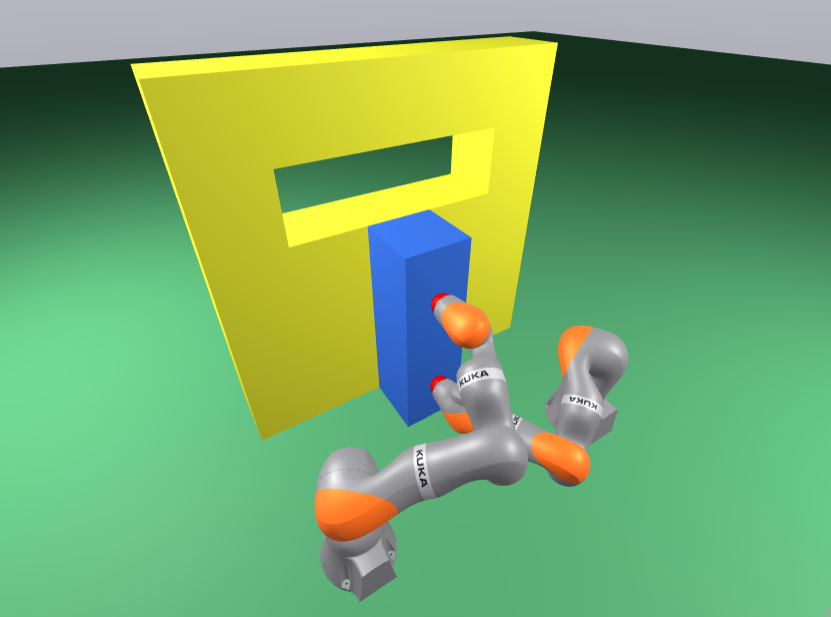}\hfill
    \includegraphics[width=0.48\columnwidth]{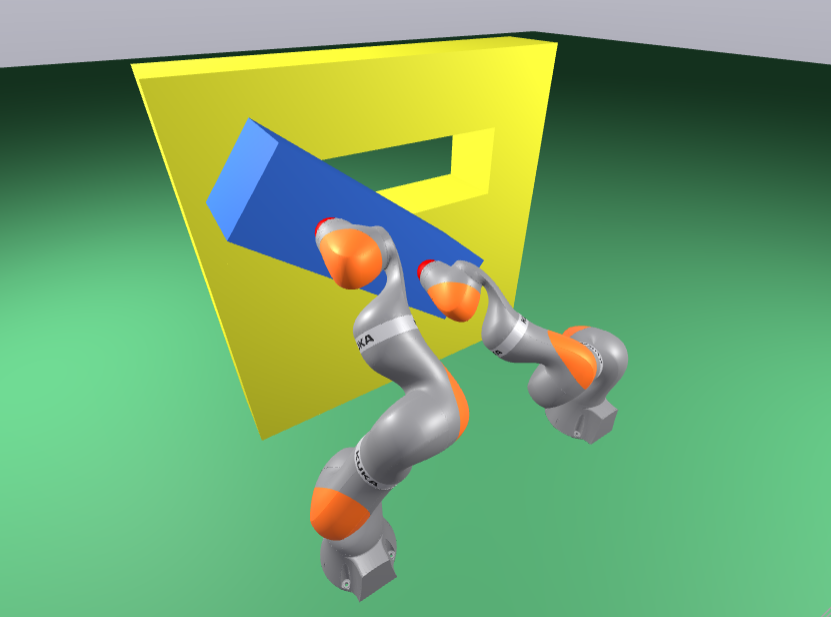}\\[1ex]
    \includegraphics[width=0.48\columnwidth]{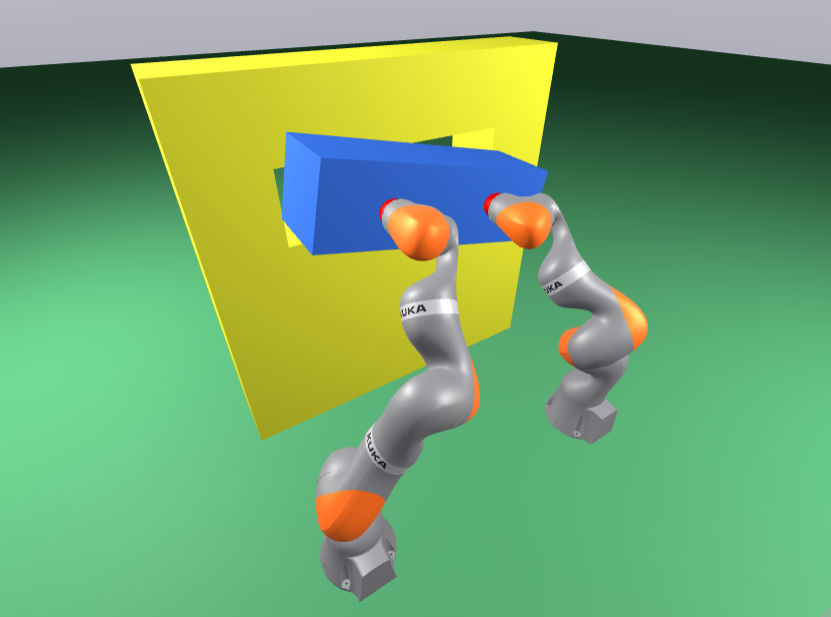}\hfill
    \includegraphics[width=0.48\columnwidth]{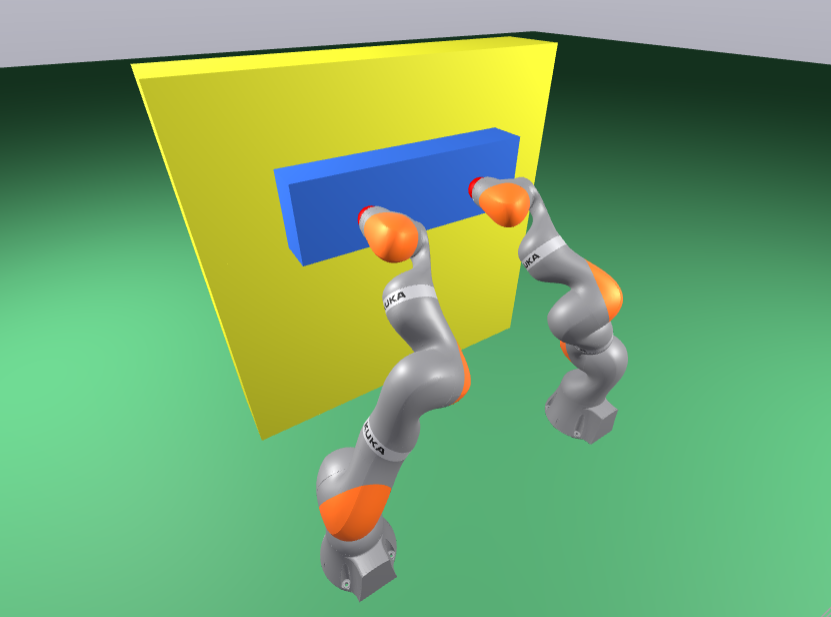}%
    \vspace{-2mm}
    \caption{Bimanual Press-Fit. The arms first press the object to the wall and rotate, and then they press-fit it into the window on the wall.}
    \label{fig:exp4_pressfit}
    \vspace{-6mm}
\end{figure}

Two iiwa7 arms that are 1~m apart manipulate a $25 \times 25 \times 80$~cm object (Fig. \ref{fig:exp4_pressfit}). The arms cooperatively press the object against the wall, slide it along the surface, and apply differential forces to rotate it into a horizontal orientation. Once horizontal, the arms cooperatively press-fit the object into a designated wall spot. This validates the effectiveness of the bimanual press-and-slide mode.

\section{DISCUSSION AND CONCLUSION}
\label{sec:conclusion}

In this paper, we presented a mode-aware framework for non-prehensile planar manipulation that mitigates the computational bottlenecks of traditional hybrid contact models. By abstracting limit surface mechanics into physically intuitive reduced-order models, we seamlessly unified single-arm and bimanual manipulation. For single-arm tasks, we introduced the body-fixed planning point, a non-holonomic stabilization point bridging car-like and unicycle-like kinematics without complex switching logic. For bimanual tasks, coordinated pushing and active Center of Pressure (CoP) steering unlock quasi-holonomic maneuverability. This formulation enables real-time, optimization-free execution via an $\mathcal{O}(1)$ algebraic force allocator, providing a highly scalable pathway for deploying complex dual-arm maneuvers on emerging bimanual and humanoid platforms.

Future work will focus on hardware validation to evaluate resilience against sensor noise, estimation delays, and unmodeled friction. Additionally, because manipulating heavy or compliant objects dynamically alters pressure distributions, we plan to incorporate the Winkler elastic foundation model for further analysis \cite{Winkler}. Integrating these contact mechanics will characterize contact patch deformation effects on the CoP, further refining our allocators for robust manipulation in unstructured environments.


\bibliographystyle{IEEEtran}
\bibliography{references}

\end{document}